# Explainable AI for Mental Health Emergency Returns: Integrating LLMs with Predictive Modeling

Abdulaziz Ahmed[1, 2*], Mohammad Saleem[1], Mohammed Alzeen[1], Badari Birur[3], Rachel E Fargason[3], Bradley G Burk[3,4], Ahmed Alhassan[3], Mohammed Ali Al-Garadi[5]

[1]Department of Health Services Administration, School of Health Professions, University of Alabama at Birmingham, Birmingham, AL United States.

[2]Department of Biomedical Informatics and Data Science, Heersink School of Medicine, University of Alabama at Birmingham, Birmingham, AL 35233, USA

[3]Department of Psychiatry and Behavioral Neurobiology, University of Alabama at Birmingham, Birmingham, AL, United States

[4]Department of Pharmacy, University of Alabama at Birmingham, Birmingham, AL, United States

[5]Department of Biomedical Informatics, School of Medicine, Vanderbilt University Medical Center, Nashville, TN, United States

## ABSTRACT

**Importance:** Emergency department (ED) returns for mental health conditions represent a significant healthcare burden. Traditional machine learning (ML) models for predicting these returns often lack interpretability for clinical implementation.

**Objective:** To evaluate whether integrating large language models (LLMs) with traditional ML approaches improves both the predictive accuracy and clinical interpretability of ED mental health returns models.

**Methods:** This retrospective cohort study analyzed 42,464 ED visits for 27,904 unique mental health patients at an Academic Medical Center in the deep South of the U.S. between January 2018 and December 2022.

**Main Outcomes and Measures:** Two outcomes were evaluated: (1) 30-day ED return prediction accuracy and (2) model interpretability through a novel LLM-enhanced explainability framework integrating SHAP (SHapley Additive exPlanations) values with contextual clinical knowledge.

**Results:** For chief complaint classification, Llama 3 (8B) with 10-shot learning outperformed traditional models, achieving 0.882 accuracy and 0.86 F1-score. In SDoH classification, LLM-based models achieved F1 scores between 0.67-0.96 for Alcohol, Tobacco, Substance Abuse, Exercise, and Home Environment. These results demonstrate the effectiveness of LLM-enhanced feature extraction in clinical prediction. The proposed ML interpretability framework leverages LLM to translate model predictions into clinically relevant explanations. LLM-extracted features improved XGBoost's AUC from 0.74 to 0.76 and AUC-PR from 0.58 to 0.61.

* **Crossponding Author**: Abdulaziz Ahmed: Email: aahmed@uab.edu; Address: 1720 2nd Ave South, Birmingham, AL, USA.

**Conclusions:** Integrating LLMs with traditional ML models yielded modest improvements in ED return prediction accuracy while substantially enhancing model interpretability. This approach offers a framework for translating complex predictive analytics into actionable clinical insights.

## 1. Introduction

Emergency department (ED) utilization for mental health (MH) conditions has reached critical levels, with significant implications for healthcare systems and patient outcomes. Currently, two-thirds of hospital ED visits annually by privately insured individuals in the U.S., 18 out of 27 million, are considered avoidable, with patients who could be treated safely and effectively in lower-cost primary care settings [1]. In emergency psychiatric services, nearly one in four patients (25.2%) return to the ED within 30 days after discharge, with 28% of these returns occurring at different facilities [2]. Psychiatric emergency rooms (PERs) are particularly overwhelmed, with ED boarding and prolonged waits for psychiatric beds reported across many regions [3].

Recent screening programs reveal that up to 17% of ED patients present with at least one unmet social need requiring immediate attention [4]. SDoH have emerged as key drivers of these utilization patterns. For instance, adults who experienced food insecurity in 2020 had 3.1 percentage points higher rates of social isolation and 9.7 percentage points higher rates of loneliness the following year compared to food-secure counterparts [5]. Community-based interventions have demonstrated potential in tackling these issues, as research indicates that a rise in MH visits at community health centers is linked to a 5% reduction in ED visits for suicidal thoughts and self-harm [6]. However, their effectiveness varies considerably depending on the type of condition. These services prove beneficial for adjustment disorders, anxiety, and mood disorders yet have a limited effect on visits associated with psychotic disorders and substance use [6]. Systemic challenges in PERs persist, including ED boarding, bed shortages, and delays in timely psychiatric assessments [2]. At the same time, critical social needs—such as housing instability, food insecurity, and social isolation—often go unaddressed in routine ED workflows [7]. Insurance coverage further complicates access, with Medicaid beneficiaries and the uninsured frequently relying on EDs as their primary source of MH care [8].

While traditional machine learning (ML) models can predict the risk of ED return using structured data, their limitations in processing unstructured clinical notes and generating interpretable outputs hinder clinical adoption [6, 9, 10]. In healthcare settings, explainability is essential—clinicians require transparent, context-aware insights to guide decision-making [11, 12]. Recent advances in LLMs offer promising avenues to address these limitations. Large Language Models (LLMs) have demonstrated the ability to process unstructured data and synthesize contextual information. This approach can enhance the interpretability of ML models by generating clinically coherent narratives that align with provider reasoning while maintaining high predictive fidelity [13-15]. Despite these innovations, the utility of LLM-enhanced frameworks to improve and explain clinical applications, particularly in clarifying ML outcomes and their related features, remains underexplored. In this study, we introduce an integrated LLM-enhanced ML framework to predict 30-day ED returns among MH patients. The system incorporates structured EHR variables, standardizes free-text SDoH inputs using few-shot prompting, and generates natural language explanations using a transformer-based LLM. This work makes the following key contributions:

- ***LLM-Augmented Feature Extraction:*** We implement few-shot learning using LLaMA 3 (8B) to classify chief complaints and harmonize non-standard SDoH text, improving feature quality for downstream modeling.
- ***Integrated Explainability Framework*:** We present a hybrid approach combining SHAP values with LLM-generated narratives contextualized by cohort-level and patient-level information to support clinical interpretation.
- ***Improved Clinical Usability:*** We demonstrate that LLM-enhanced features yield consistent gains in predictive accuracy and significantly improve interpretability, an essential step toward real-world adoption of AI in psychiatric ED care.

**Statement of Significance**

| Problem or Issue | What is Already Known | What this Paper Adds | Who Would Benefit from the Knowledge in this Paper |
|---|---|---|---|
| ED returns for MH conditions are common and burdensome, yet existing predictive models lack clinical interpretability. | Traditional ML models can predict ED returns using structured data but fail to leverage unstructured clinical narratives and often lack explainability needed for adoption. | This study introduces a layered framework combining ML, SHAP, and LLMs to enhance prediction accuracy and generate narrative explanations contextualized with clinical reasoning and population statistics. | Clinicians, informatics researchers, and hospital decision-makers aiming to understand and reduce MH-related ED returns using interpretable AI tools. |

## 2. Related Work

### 2.1 Emergency Department Returns and MH Utilization

EDs serve as a critical entry point for individuals experiencing acute MH crises, yet they are often ill-equipped to provide comprehensive psychiatric care [16]. Data from the National Hospital Ambulatory Medical Care Survey indicate that adults with MH disorders accounted for 52.9 ED visits per 1,000 adults annually from 2017 to 2019, with higher rates among younger adults and those covered by Medicaid [17]. Notably, patients with MH conditions often experience longer ED stays—over 40% of visits by adults with MH disorders lasted four hours or more, compared to about 25% among those without such disorders, and this disparity increased with age [18]. Extended stays reflect both the complexity of psychiatric assessments and the limited availability of inpatient psychiatric beds, exacerbating ED boarding and straining emergency resources [16].

Patients with severe mental illnesses, such as schizophrenia, bipolar disorder, and substance use disorders, are particularly vulnerable to frequent ED utilization. These individuals often face barriers to accessing outpatient MH services, including stigma, lack of transportation, and insufficient community-based support [16, 19]. Moreover, SDoH such as housing instability, unemployment, social isolation, and food insecurity are strongly associated with increased ED visits for MH reasons For example, a large study in California found that patients with MH diagnoses who were also homeless or had co-occurring substance use disorders were significantly more likely to be frequent ED users (defined as more than four visits in a year). Food insecurity, in particular, has been linked to loneliness and heightened MH crisis presentation [19].

While community-based programs have shown effectiveness in reducing ED revisits for conditions like anxiety and depression, their impact is limited for more complex MH conditions such as psychosis and substance use disorders [19, 20]. These limitations are compounded by persistent systemic barriers—including insufficient psychiatric beds, delayed assessments, and care fragmentation—which hinder timely intervention and follow-up [16]. Medicaid recipients and uninsured individuals are especially dependent on EDs for MH care, revealing a deeper intersection of healthcare access, insurance coverage, and socioeconomic vulnerability [18, 20]

## 2.2 ML in Predicting ED Returns

Predicting emergency department (ED) returns, particularly for mental health (MH) patients, has gained increasing attention due to the high rates of unscheduled revisits and the considerable burden they impose on healthcare systems. Traditional machine learning (ML) models—such as logistic regression, XGBoost, and random forests—have been widely used to analyze structured electronic health record (EHR) data, including demographics, clinical diagnoses, and prior ED utilization, demonstrating moderate predictive success [10, 21]. However, these models often lack interpretability, limiting their clinical applicability [10, 21]. A key limitation is their dependence on structured data, which excludes the rich contextual information found in unstructured narratives. Clinical notes, triage documentation, and discharge summaries often contain vital insights into patient behavior, social context, and provider reasoning—elements particularly relevant for MH assessments. Recent studies underscore the value of integrating both structured and unstructured data to enhance prediction accuracy and clinical relevance in ED return models [22, 23].

## 2.3 Explainable AI (XAI) in Healthcare

Explainability is increasingly recognized as a prerequisite for the adoption of AI in clinical settings, as it enables clinicians to understand, trust, and act upon model predictions [10]. SHapley Additive exPlanations (SHAP) is a popular method for feature attribution, providing insight into which variables most influence model outputs [24]. However, SHAP values alone often lack the narrative context necessary for actionable clinical decision-making, especially in the complex and nuanced domain of MH [25]. Recent advances in explainable AI (XAI) have sought to bridge this gap by integrating SHAP values with clinical narratives or domain knowledge, thereby enhancing both the interpretability and clinical relevance of model outputs [26]. For example, some approaches use LLM-generated textual explanations in conjunction with SHAP to provide clinicians with clear, context-rich rationales for model predictions. However, these solutions often require manual rule creation or domain-specific templates, limiting their scalability and generalizability [27].

### 2.4 LLMs for Clinical Data Processing

Recent advancements in LLMs, such as ClinicalBERT, BlueBERT, ChatGPT, and LLaMA, have revolutionized natural language processing (NLP) in healthcare [28-30]. These models excel at processing unstructured clinical data—such as clinical notes, discharge summaries, and triage narratives—which traditional ML approaches struggle to utilize effectively. LLMs can automatically extract features, standardize ambiguous clinical language, and generate high-quality representations for downstream predictive tasks [28-30]. Few-shot learning techniques have further enhanced the utility of LLMs in clinical settings, enabling accurate classification of chief complaints and standardization of clinical narratives with limited labeled data. Our study significantly advances the existing literature by integrating LLMs to enhance both predictive accuracy and clinical interpretability for ED returns among MH patients. **Table 1** summarizes the key differences and highlights our study's unique contributions compared to previous research.

***Table 1: Key Differences Between Previous Studies and Our study***

| Aspect | Previous Studies | Our Study Contributions |
|---|---|---|
| Data Utilization | Primarily structured data | Integrated structured and unstructured data with LLM |
| Feature Extraction | Traditional methods | LLM-based few-shot learning (Accuracy: 0.882, F1-score: 0.86) |
| Explainability | Numeric SHAP values | Clinically coherent, LLM-enhanced narratives |
| SDoH Standardization | SDoH Standardization | Automated and accurate LLM-based extraction |

## 3. Research Methodology

### 3.1 Research Framework

Figure 1 presents the overall framework of our proposed approach, which integrates LLMs with traditional ML methods to enhance both prediction accuracy and interpretability for ED returns among MH patients. The framework begins with the extraction of both structured and unstructured data from the electronic health record, including demographics, clinical measures (e.g., vital signs, ICD codes, Emergency Severity Index [ESI]), visit-related information, and unstructured text fields such as chief complaints and social determinants of health (SDoH). Structured data undergo preprocessing, including categorical harmonization, binning of temporal and clinical variables, and imputation of missing values. The unstructured text is processed using LLaMA 3 (8B), a transformer-based LLM, which classifies chief complaints into clinically meaningful categories (e.g., Pain, Psychiatric, Injury) and extracts structured representations of SDoH features such as alcohol use, housing status, and exercise habits. These enriched features are then combined with structured variables and used to train multiple ML algorithms, including XGBoost, neural networks, and gradient boosting, with model performance evaluated using standard metrics such as accuracy, F1-score, AUC, and AUC-PR. To address class imbalance, random oversampling is applied to the training set. The final component of the framework is an explainability module that integrates SHAP (SHapley Additive exPlanations) values with contextual information such as individual patient attributes and population-level statistics. These are synthesized by the LLM into natural language explanations that provide actionable, patient-specific insights. The output is an interpretable ML model capable of not only predicting 30-day ED return risk but also communicating its predictions in a manner that aligns with clinical reasoning and supports decision-making.

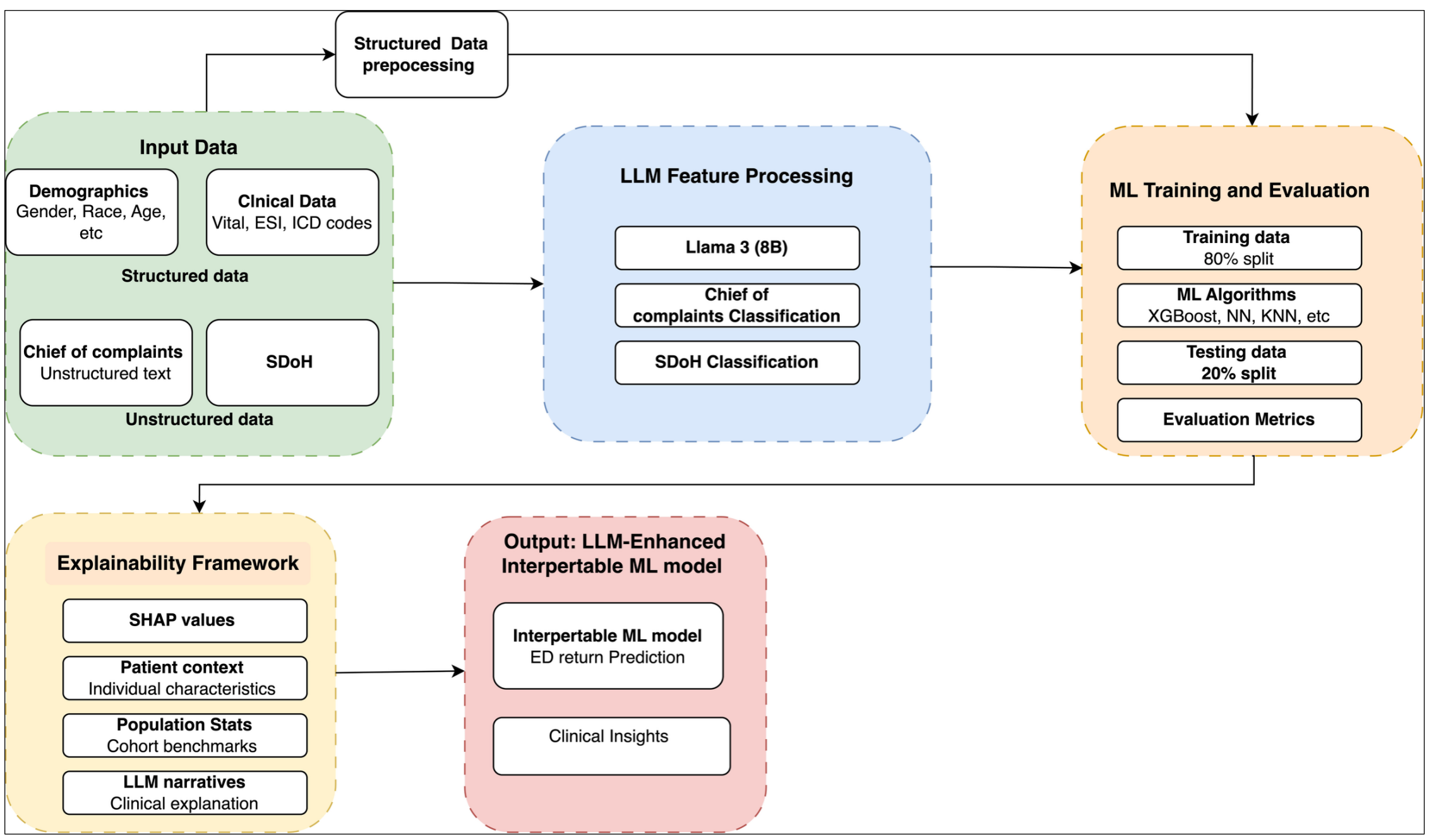

***Figure 1. Proposed framework.***

### 3.2 Data Acquisition and Preprocessing

This retrospective cohort study analyzed structured electronic health record (EHR) data from an academic medical center in the Deep South of the United States, covering ED visits between January 2018 and December 2022. The final cohort included 42,464 visits from 27,904 unique adult patients diagnosed with mental and behavioral disorders, identified using ICD-10-CM codes beginning with "F." [31] Patients under 18 years of age or without qualifying MH diagnoses were excluded.

Structured data were extracted from multiple sources, including demographic variables (e.g., gender, race, ethnicity, age), clinical measurements (e.g., systolic and diastolic blood pressure, heart rate, respiratory rate, oxygen saturation, and temperature), insurance status, Emergency Severity Index (ESI), visit timing (e.g., hour of day, weekend vs. weekday), and return visit history. Diagnoses were also included as standardized ICD-10 codes. All structured variables underwent a rigorous preprocessing pipeline. Multiple vital sign recordings per visit were averaged to generate representative values. Continuous features were standardized using z-score normalization. Clinical variables such as blood pressure, temperature, heart rate, and BMI were binned into standard clinical categories to enhance interpretability and support risk stratification. Age was grouped into four clinically relevant categories: 18–30, 31–45, 46–60, and over 60 years. Temporal variables were transformed into categorical indicators, including time-of-day intervals and weekend vs. weekday presentation. The summary statistics for the collected features are shown in Table 2.

Missing data were handled using variable-specific strategies. Continuous variables with missing values were imputed using K-Nearest Neighbors (KNN) imputation.[32] Categorical variables with more than 20% missingness were excluded from further analysis. For the remaining categorical features, missing values were encoded as "Unknown" to preserve completeness.

The primary outcome variable was defined as a binary indicator of whether a patient returned to the ED within 30 days of the index visit. This structured preprocessing approach provided a clean and analytically robust dataset for subsequent model development and evaluation. By combining tailored imputation, principled exclusion criteria, and appropriate encoding strategies, we ensured the creation of a clean, consistent, and analytically robust structured dataset. This preprocessing pipeline laid the foundation for more reliable model training, evaluation, and interpretation.

***Table 2. Study Population Characteristics, Including SDoH, Demographic, Clinical, and Visit-Related Features***

| **Features** | **Ranges for Date/Time Features, Average ± Standard Deviation for Numerical Features, % for Categorical Features** |
|---|---|
| **Number Visits Past 2 Months** | 1.03 ± 2.75 (0.0–52.0) |
| **Gender** | M: 55.06%; F: 44.94% |
| **Marital Status** | Single: 63.07%; Married: 17.79%; Divorced: 9.78%; Widowed: 3.89%; Unknown: 3.17%; Separated: 2.09%; Life Partner: 0.21% |
| **Race** | White: 50.32%; Black or African American: 45.57%; Other: 2.38%; Decline/Refuse: 1.25%; Unknown: 0.48% |
| **Ethnic Group** | Non-Hispanic/Latino: 95.20%; Unknown: 1.98%; Not Reported: 1.69%; Hispanic/Latino: 1.07%; Multiple: 0.06% |
| **Language** | English: 96.66%; Other: 3.33%; Sign Language: 0.01% |
| **Insurance** | Government Insurance: 34.47%; Self-Pay: 33.74%; Private Insurance: 22.71%; Other: 9.08% |
| **ESI Level** | 3: 48.13%; 2: 27.68%; 4: 20.46%; 5: 2.93%; 1: 0.80% |
| **Month of Year, Day of Month, Hour of Day** | 1-12 Months, 1-31 Days, 1-24 Hours |
| **Weekend** | False: 73.30%; True: 26.70% |
| | 9: 8.64%; 8: 8.58%; 6: 8.57%; 7: 8.56%; ... |
| **Returned in 30 Days** | 0.0: 73.40%; 1.0: 26.60% |
| **Systolic Blood Pressure** | Elevated: 37.92%; Hypertension: 33.51%; Normal: 28.14%; Low: 0.44% |
| **Diastolic Blood Pressure** | Normal: 41.98%; Elevated: 29.27%; Hypertension: 24.36%; Low: 4.40% |
| **Temperature** | Normal: 95.64%; Fever: 2.98%; Below Normal: 1.27%; Hypothermia: 0.12% |
| **Heart Rate** | Normal: 83.10%; Tachycardia: 14.62%; Bradycardia: 2.28% |
| **Age** | 31_45: 38.17%; 18_30: 26.46%; 46_60: 22.76%; Over_60: 12.61% |
| **BMI** | Normal Weight: 38.39%; Overweight: 29.00%; Obese: 28.86%; Underweight: 3.75% |
| **Chief Complaint** | Pain: 45.82%; Psychiatric: 36.25%; Injury: 9.32%; Infection: 8.15%; Unclear: 0.46% |
| **Tobacco Use** | Current Use: 35.52%; Unclear/Other: 34.07%; No Use: 21.39%; Former Use: 8.05%; Occasional Use: 0.89%; Prescribed Use: 0.08% |
| **Nutrition Health** | Unclear/Other: 79.64%; Moderate Nutrition: 10.75%; Good Nutrition: 4.51%; Poor Nutrition: 2.73%; Special Diet: 1.30%; Assistance Required: 1.06% |
| **Home Environment** | Unclear/Other: 69.02%; Independent: 16.12%; Family Support: 8.83%; Homeless: 3.21%; Living with Friends: 1.66%; Assisted Living: 0.75%; Unstable Housing: 0.40% |
| **Alcohol Use** | Unclear/Other: 35.40%; No Alcohol Use: 31.27%; Current Alcohol Use: 17.39%; Past Alcohol Use: 8.19%; Occasional Use: 7.58%; Recovering: 0.16% |
| **Exercise** | Unclear/Other: 60.35%; No Exercise: 30.89%; Light Exercise: 5.50%; Moderate Exercise: 2.80%; Vigorous Exercise: 0.39%; Physical Therapy: 0.08% |
| **Sexual Orientation** | Unclear/Other: 91.89%; Heterosexual: 5.57%; Gender Non-Binary: 1.75%; Homosexual: 0.43%; Transgender: 0.17%; Bisexual: 0.16%; Asexual: 0.01%; Queer/Other: 0.01% |
| **Substance Abuse** | No Use: 38.88%; Unclear/Other: 33.48%; Recreational Use: 10.59%; Current Use: 10.23%; Former Use: 5.74%; Prescribed Use: 1.07% |

## 3.3 Chief Complaint Classification Methodology

To convert free-text chief complaints into structured, clinically meaningful categories suitable for downstream predictive modeling, we implemented and compared four classification strategies: (1) traditional ML models using bag-of-words and TF-IDF features, (2) transformer-based contextual embeddings using Clinical Longformer, (3) domain-specific fine-tuning with BlueBERT, and (4) few-shot classification using a LLM (LLaMA 3). Each method classified chief complaints into one of five categories—Pain, Psychiatric, Injury, Infection, and Unclear—following the classification scheme proposed by Kuykendal et al. [33]. All methods were applied to the same annotated dataset using consistent training (70%), validation (20%), and test (10%) splits.

### *3.3.1 Traditional ML with Bag-of-Words and TF-IDF*

We first applied traditional text classification techniques using two standard feature extraction methods: Count Vectorizer (bag-of-words) and Term Frequency–Inverse Document

Frequency (TF-IDF).[34] After standard text preprocessing (e.g., lowercasing, punctuation removal), each chief complaint was transformed into a high-dimensional sparse vector, retaining the top 5,000 features. These feature matrices were used to train and evaluate three commonly used classifiers—XGBoost,[35] Random Forest,[36] and Support Vector Machine (SVM)[37]. These models were implemented using scikit-learn[38] and XGBoost libraries[39] and evaluated using standard multi-class metrics.

### *3.3.2 Transformer-Based Embedding with Clinical Longformer*

To capture richer contextual information, we employed the Clinical Longformer, a transformer-based model pretrained on long-form clinical texts. [40] Each chief complaint was tokenized and passed through the model, and the final-layer hidden state corresponding to the [CLS] token was extracted as a dense feature vector. These embeddings were used as input to XGBoost, Random Forest, and SVM classifiers, using the same data partitions and evaluation metrics described above. The model was selected for its ability to process lengthy clinical narratives, making it particularly well-suited for variable-length ED documentation. Embedding generation followed the standard practice of pooling contextual vectors from the final transformer layer.[41]

### *3.3.3 Fine-Tuned Domain Model: BlueBERT*

Next, we fine-tuned BlueBERT, a domain-specific variant of BERT pretrained on PubMed abstracts and MIMIC-III clinical notes.[42] Chief complaints were tokenized using the BlueBERT tokenizer (maximum length: 512 tokens), and datasets were constructed using Hugging Face's datasets API. Fine-tuning was performed for 20 epochs using the Trainer API with AdamW optimization [43], a batch size of 16, and evaluation at each epoch. The final model produced classification logits over the five target categories, and the predicted label was selected based on the highest softmax score. BlueBERT's pretraining on biomedical texts made it particularly effective for capturing domain-specific terminology and abbreviations frequently used in ED notes.

### *3.3.4 Few-Shot Learning with LLM (LLaMA 3)*

Lastly, we used LLaMA 3 (8B) in a few-shot classification setup via the LangChain framework and local Ollama deployment [44]. This method did not require fine-tuning; instead, we constructed prompts containing 10 representative examples from the training data, each consisting of a chief complaint, major category, and subcategory (see Appendix A.1 for the prompt template). The model was asked to classify a new complaint and return both the major and subcategory without additional explanation. Subcategories were drawn from a comprehensive list based on expert-defined clinical themes (e.g., “Chest pain,” “Fall,” “Respiratory infection symptoms”). The few-shot paradigm allowed the LLM to leverage in-context learning to classify complaints based on semantic similarity to the examples, an approach increasingly adopted for healthcare NLP tasks.

### 3.4 Social Determinants of Health (SDoH) Classification Using LLM

Several social determinants of health (SDoH) fields—such as alcohol use, nutrition, tobacco use, substance use, exercise, housing environment, and sexual orientation—were available in the EHR as structured columns, yet their values were expressed in highly heterogeneous and ambiguous free-text formats. These entries often included idiosyncratic phrasing, abbreviations, or unstructured narrative inputs that lacked standardized coding, making them unsuitable for direct inclusion in downstream models. To address this, we employed a LLM (LLaMA 3, 8B), deployed locally via the Ollama framework, to standardize these fields through few-shot classification. For each domain, we predefined a discrete set of clinically meaningful category labels based on expert knowledge and guidelines (e.g., for alcohol use: "No Alcohol Use," "Current Alcohol Use," "Past Alcohol Use," "Occasional Use," "Recovering," and "Unclear/Other"), and constructed domain-specific prompt templates containing natural language instructions and multiple representative examples drawn from real-world entries (see Appendix A.2-A.8 for the complete prompt templates). These prompts guided the LLM to map each nonstandard value to its appropriate standardized category using semantic similarity and contextual alignment, without requiring explicit fine-tuning. The classification pipeline was implemented in Python using the LangChain interface. For each patient entry, the model's predicted label was stored as a new standardized categorical variable. This methodology enabled efficient and reproducible transformation of noisy SDoH fields into structured representations suitable for statistical analysis and predictive modeling, while leveraging the flexibility and generalization capacity of few-shot in-context learning with LLMs.

### 3.5 Predictive Modeling

We developed a comprehensive ML framework to predict the risk of ED return among patients presenting with MH conditions. The framework integrated EHR variables—including demographic, clinical, and encounter-level information—with SDoH features derived from structured fields and harmonized using LLM classification. To evaluate the added value of enriched features, we constructed two comparative datasets: one containing all available features (baseline + LLM-processed chief complaint and SDoH), and another excluding SDoH variables to isolate the contribution of socioeconomic indicators.

Model development focused on five supervised learning algorithms: Logistic Regression, Neural Network (Multilayer Perceptron), Adaptive Boosting (AdaBoost), Gradient Boosting, and eXtreme Gradient Boosting (XGBoost). All classifiers were implemented using the scikit-learn and xgboost Python libraries. For each model, hyperparameter optimization was conducted via GridSearchCV with 3-fold cross-validation on the training set to maximize performance. Key hyperparameters such as learning rate, number of estimators, regularization strength, and network architecture (for Neural Network) were systematically tuned. To address class imbalance in the outcome (i.e., ED return vs. no return), random oversampling was applied exclusively to the training set, ensuring that the minority class was adequately represented without data leakage into the test set.

The dataset was split into 80% training and 20% testing partitions. Predictive performance was assessed using standard classification metrics, including accuracy, sensitivity, specificity, F1 score, and the area under the receiver operating characteristic curve (AUC). Sensitivity and specificity offered insight into the models' ability to correctly identify patients at risk versus those not at risk, while the F1 score provided a balanced measure of precision and recall. AUC was used as a global indicator of discriminative performance. All models were trained and evaluated under consistent data partitions and preprocessing protocols to ensure methodological rigor and comparability. By training parallel models with and without LLM-derived SDoH variables, we were able to quantify the incremental predictive value of incorporating socioeconomic context into risk stratification for MH-related ED returns.

### 3.6 Enhancing Explainability Framework with an LLM

To enhance the interpretability of ML predictions in clinical settings, we employ an explainability framework that integrates LLMs with patients-specific information (Figure 2). This approach combines feature-level attributions with contextual background information, resulting in rich, clinically meaningful narratives that align with the reasoning patterns used by healthcare professionals. Data from the study cohort—including SDoH variables, structured features, and LLM-derived attributes—are utilized in the development and testing of ML models. This process culminates in an explainability step, which integrates SHAP values [45] and patient-specific information to produce interpretable outputs such as cohort statistics, SHAP visualizations, and patient-centered narratives. By incorporating SHAP values, we can assess each feature's contribution to a patient's predicted risk, providing granular, quantitative insights into feature importance. However, the numerical nature of SHAP values often limits clinical interpretability. To bridge this gap, we leverage a domain-specific knowledge repository that includes population-level cohort statistics, risk factor ranges derived from the ML model, and individual patient characteristics. The LLM synthesizes the SHAP values and the retrieved context into cohesive narratives that reflect real-world clinical reasoning, translating the raw output of the ML models into understandable terms (see Appendix A.9 for the prompt template used in generating these explanations). This enables clinicians to comprehend the model's predictions in actionable terms, enhancing the transparency and trustworthiness of the predictions. We detail the components of our explainability framework as follows:

A. **Deriving SHAP-Based Feature Attributions:** The first step in enhancing explainability involves training a predictive ML model and calculating SHAP values to assess each feature's contribution to a patient's predicted risk of an ED visit. SHAP values provide granular, quantitative insights into feature importance, but their numerical nature often limits clinical interpretability.
B. **Contextualization Through Document Retrieval:** To bridge the gap between SHAP outputs and clinical actionability, we leverage a domain-specific knowledge repository. This repository includes population-level cohort statistics, risk factor ranges derived from the model, and individual patient characteristics (input features used in the predictive model).
C. **Generating Clinically Coherent Narratives:** The LLM then synthesizes the SHAP values and the retrieved domain-specific context into a cohesive narrative that reflects real-world clinical reasoning. These narratives translate the raw output of the ML models into understandable terms, linking patient attributes—such as acuity level, time-of-day presentation, and other risk factors—to established medical knowledge. Thus, clinicians can understand the model's predictions in actionable terms. As illustrated in Figure 1, the explainability framework aligns patient-specific attributes with population benchmarks and temporal patterns. A low-risk patient may exhibit presentation times and acuity levels consistent with population norms, suggesting no significant deviation from baseline risk. Conversely, a high-risk patient may display temporal patterns or acuity levels linked to acute exacerbations, providing insights into the factors driving their elevated risk.

**D. Assessment Protocol for Explainability Framework Reliability:** The reliability of LLM-generated clinical explanations was evaluated through a structured assessment protocol. All explanations underwent systematic cross-referencing against three data sources: source patient records, retrieved reference documents, and population-level statistics. We assessed four dimensions: factual accuracy (numerical values, temporal relationships), clinical consistency (alignment with medical knowledge), logical coherence (internal consistency), and feature attribution accuracy (correspondence with SHAP values). The potential for hallucinations—fabricated or unsupported information—was monitored throughout the evaluation. A severity classification system categorized errors as minor (no clinical impact), moderate (potential interpretation issues), or severe (impact on clinical decision-making). Two experts independently reviewed all explanations for potential errors, hallucinations, and clinical significance.

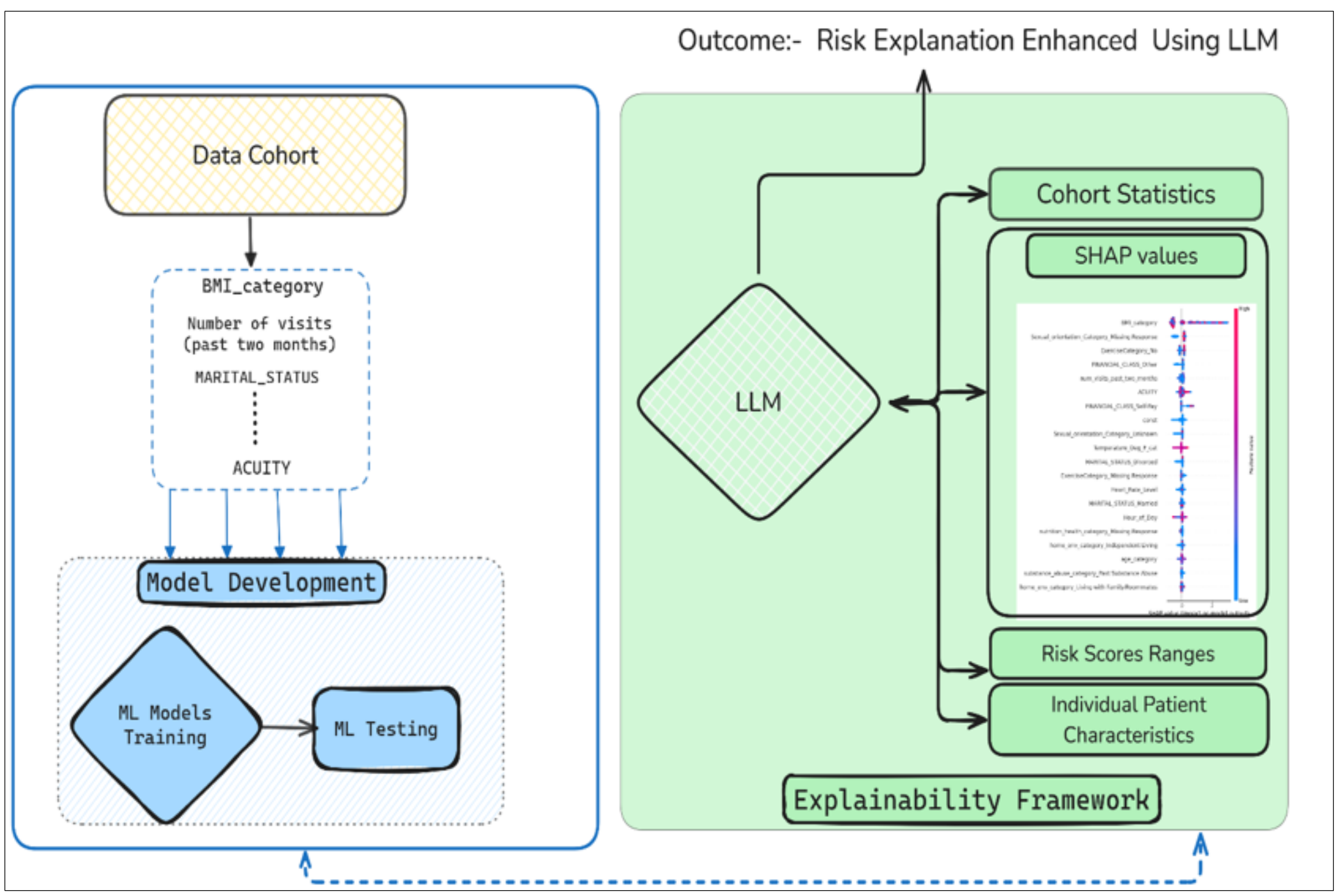

*Figure 2: Integration of LLMs to Explainability Framework for ED Return Risk*

## 4. Results

### 4.1 LLM features extraction performance results

This section evaluates the performance of the LLM (Llama 3:8-billion) in feature extraction for chief complaint and SDoH classifications. Few-shot learning approaches are compared to traditional ML and pre-trained models.

#### *4.1.1 Chief Complaint Classification*

The classification of chief complaints was evaluated using traditional ML models, pre-trained language models, and few-shot learning approaches. Among these, the LLM (Llama 3, 8-billion) with 10-shot learning demonstrated the best performance across all metrics, achieving an Accuracy of 0.882, Precision of 0.95, Recall of 0.88, and an F1-Score of 0.86 (Table 3). This significantly outperformed traditional models like XGBoost (Accuracy: 0.59, F1-Score: 0.53) and pre-trained models such as BlueBERT (Accuracy: 0.63, F1-Score: 0.59). Other few-shot configurations, including 5-shot (Accuracy: 0.816) and 20-shot (Accuracy: 0.803), also performed well but were slightly less effective than the 10-shot setting.

***Table 3: Performance Metrics for Chief Complaint Classification Using Different Models***

| **Model** | **Accuracy** | **Precision** | **Recall** | **F1-Score** |
|---|---|---|---|---|
| **XGBoost** | 0.59 | 0.48 | 0.59 | 0.53 |
| **Random Forest** | 0.59 | 0.44 | 0.59 | 0.50 |
| **SVM** | 0.62 | 0.41 | 0.62 | 0.50 |
| **BlueBERT** | 0.63 | 0.56 | 0.63 | 0.59 |
| **Llama 3( 8-billion) -Few-shot (20)** | 0.803 | 0.88 | 0.80 | 0.75 |
| **Llama 3( 8-billion)-Few-shot (5)** | 0.816 | 0.91 | 0.81 | 0.77 |
| **Llama 3( 8-billion)- Few-shot (10)** | 0.882 | 0.95 | 0.88 | 0.86 |

#### *4.1.2 SDoH Classification*

The LLM (Llama 3, 8-billion) with 10-shot learning achieved strong performance across SDoH categories, particularly in Alcohol, Tobacco, and Substance Abuse, with an overall Accuracy of 0.95 and a weighted F1-Score of 0.96. Sensitivity ranged from 0.63 (Home Environment) to 0.95 (Alcohol and Tobacco), while Specificity remained consistently high (0.94–0.99). The model performed best in Alcohol, Tobacco, and Substance Abuse (F1: 0.96–0.89) but showed moderate performance in Sexual Orientation and Nutrition (F1: 0.79–0.72) and lower in Exercise and Home Environment (F1: 0.70–0.67). These results highlight its reliable classification across diverse and challenging variables (Table 4).

*Table 4. Performance Metrics for SDoH Classification Using LLM (Llama 3, 8-billion) with 10-Shot Learning*

| Category | Accuracy | Precision (Weighted) | Sensitivity/Recall (Weighted) | F1 Score (Weighted) |
|---|---|---|---|---|
| **Alcohol** | 0.95 | 0.99 | 0.95 | 0.96 |
| **Exercise** | 0.70 | 0.74 | 0.70 | 0.70 |
| **Home_Environment** | 0.63 | 0.78 | 0.63 | 0.67 |
| **Nutrition** | 0.68 | 0.89 | 0.68 | 0.72 |
| **Sexual_Orientation** | 0.75 | 0.90 | 0.75 | 0.79 |
| **Substance_Abuse** | 0.85 | 0.99 | 0.85 | 0.89 |
| **Tobacco** | 0.95 | 0.99 | 0.95 | 0.96 |

## 4.2 Predicative Models for ER MH Return Visits: ML without/with LLM Features Extractions

This section evaluates the performance of predictive models for ED mental and behavioral health return visits using two distinct approaches: (1) ML models trained on traditional features alone and (2) ML models enhanced with features extracted using large lLLMs. Performance metrics, including Accuracy, Precision, Recall, F1-Score, and the AUC, were used to assess the predictive capability of each approach. The results demonstrate that including LLM-extracted features consistently improved model performance across multiple metrics.

### *4.2.1 Performance of Models Without LLM Feature Extraction*

Table 5 presents the performance metrics for models trained exclusively on traditional features. Neural Network, AdaBoost, Gradient Boosting, and XGBoost all achieved the highest accuracy (0.79), with Gradient Boosting and XGBoost exhibiting the highest precision (0.72). Among them, Neural Network had the highest F1-score (0.47), while Gradient Boosting and XGBoost followed closely (0.45). The AUC values ranged from 0.68 (Logistic Regression) to 0.75 (Gradient Boosting), indicating moderate discriminative ability. In terms of AUC-PR, Neural Network had a score of 0.57, while AdaBoost, Gradient Boosting, and XGBoost achieved the highest scores (0.58). Logistic Regression showed the weakest performance across all metrics, with the lowest recall (0.31), F1-score (0.41), AUC (0.68), and AUC-PR (0.51), suggesting it struggled more in distinguishing positive cases effectively.

### *4.2.2 Performance of Models with LLM Feature Extraction*

Table 5 highlights the performance of models enhanced with LLM-extracted features, leading to noticeable improvements in key metrics. XGBoost, AdaBoost, and Gradient Boosting achieved the highest AUC (0.76), while Neural Network improved slightly to 0.75. The addition of LLM features resulted in higher precision, recall, and AUC-PR values for most models. Neural

Network, for example, maintained its F1-score of 0.47 but improved in precision (0.71) and AUC-PR (0.59). Similarly, AdaBoost and Gradient Boosting saw an increase in AUC-PR to 0.60, reflecting better overall classification performance. XGBoost remained strong, improving in recall (0.34) and F1-score (0.46), while achieving the highest AUC-PR (0.61) along with AdaBoost and Gradient Boosting. Logistic Regression, though slightly improving in AUC (0.70) and AUC-PR (0.54), continued to underperform compared to other models, reinforcing its weaker ability to capture complex patterns.

***Table 5: Models' Performance.***

| | **Model** | **Accuracy** | **Precision** | **Recall** | **F1_Score** | **AUC** | **AUC-PR** |
|---|---|---|---|---|---|---|---|
| ***Performance without LLM extraction*** | NeuralNetwork | 0.79 | 0.69 | 0.36 | 0.47 | 0.74 | 0.57 |
| | AdaBoost | 0.79 | 0.70 | 0.34 | 0.46 | 0.74 | 0.58 |
| | LogisticRegression | 0.77 | 0.65 | 0.31 | 0.41 | 0.68 | 0.51 |
| | GradientBoosting | 0.79 | 0.72 | 0.32 | 0.45 | 0.75 | 0.58 |
| | XGBoost | 0.79 | 0.72 | 0.32 | 0.45 | 0.74 | 0.58 |
| ***Performance with adding LLM feature extractions*** | NeuralNetwork | 0.79 | 0.71 | 0.35 | 0.47 | 0.75 | 0.59 |
| | AdaBoost | 0.79 | 0.71 | 0.35 | 0.46 | 0.76 | 0.60 |
| | LogisticRegression | 0.78 | 0.68 | 0.30 | 0.42 | 0.70 | 0.54 |
| | GradientBoosting | 0.79 | 0.71 | 0.34 | 0.46 | 0.76 | 0.60 |
| | XGBoost | 0.79 | 0.72 | 0.34 | 0.46 | 0.76 | 0.61 |

## 4.3 Explainability results

### *4.3.1 Clinical Validation of LLM-Generated Explanations*

In analyzing 100 randomly selected explanations, 99 demonstrated complete alignment across all assessment dimensions. A single explanation contained one numerical discrepancy (reporting a risk factor as 92 instead of 93), classified as a minor error with no clinical significance. All explanations maintained clinical validity and showed complete concordance with source documentation and SHAP-derived feature rankings. Independent expert reviews confirmed the absence of moderate or severe errors that could affect clinical interpretation or decision-making. The observed error rate was 1% (1/100), comprising solely the single minor numerical discrepancy.

### *4.3.2 Comparative Analysis of SHAP and Explainability Framework*

Figure 2 shows the SHAP summary plot illustrates the most influential features contributing to the model's predictions of MH emergency return risk. Each feature is plotted based on its SHAP value, which indicates its impact on the model's output. Features toward the top of the graph are the most impactful. Positive SHAP values (toward the right) push the prediction toward high risk, while negative SHAP values (toward the left) suggest lower risk. The color gradient represents the actual value of the feature: red indicates a high value, and blue a low value. Among the most significant predictors is the number of visits in the past two months, where higher values are strongly associated with increased risk. Features such as elevated heart rate (tachycardia) also contribute to higher risk. On the other hand, characteristics like having private insurance, being female, and being married generally reduce risk. Several social determinants, including exercise behavior, substance abuse, and housing status (e.g., being homeless or having an unclear home environment), also demonstrate meaningful influence on the model's risk classification. Importantly, categorical features marked as "Unclear/Other" (e.g., in exercise, substance use) can contribute variably, potentially reflecting missing data or ambiguous health profiles. Overall, the plot underscores the importance of both clinical indicators and social factors in shaping MH return risk predictions.

Table 6 presents a side-by-side comparison of explainability outputs for two patients—one classified as high risk and the other as low risk for a MH emergency return—using a SHAP summary bar plot and a corresponding LLM-generated narrative explanation. The SHAP bar plot visually highlights the top features influencing the model's prediction, ranked by SHAP value magnitude, while the LLM transforms this data into a plain-language narrative using the SHAP values and corresponding population statistics as input. Importantly, the LLM does not generate or infer new insights—it simply rephrases the SHAP outputs to support clinical interpretation. In the high-risk case, both the SHAP plot and the LLM explanation identify the same top contributing features, such as frequent visits in the past two months and elevated heart rate, reinforcing the model's rationale. The LLM narrative further contextualizes these features by comparing the patient's values to population averages, aiding interpretability. In the low-risk example, the SHAP plot displays lower-magnitude feature contributions, which the LLM mirrors in a concise explanation emphasizing the lack of strong risk indicators. Together, the SHAP graph and LLM-based explanation serve complementary roles—while the SHAP graph quantifies feature impact, the LLM provides a narrative summary to enhance clarity and accessibility for clinicians.

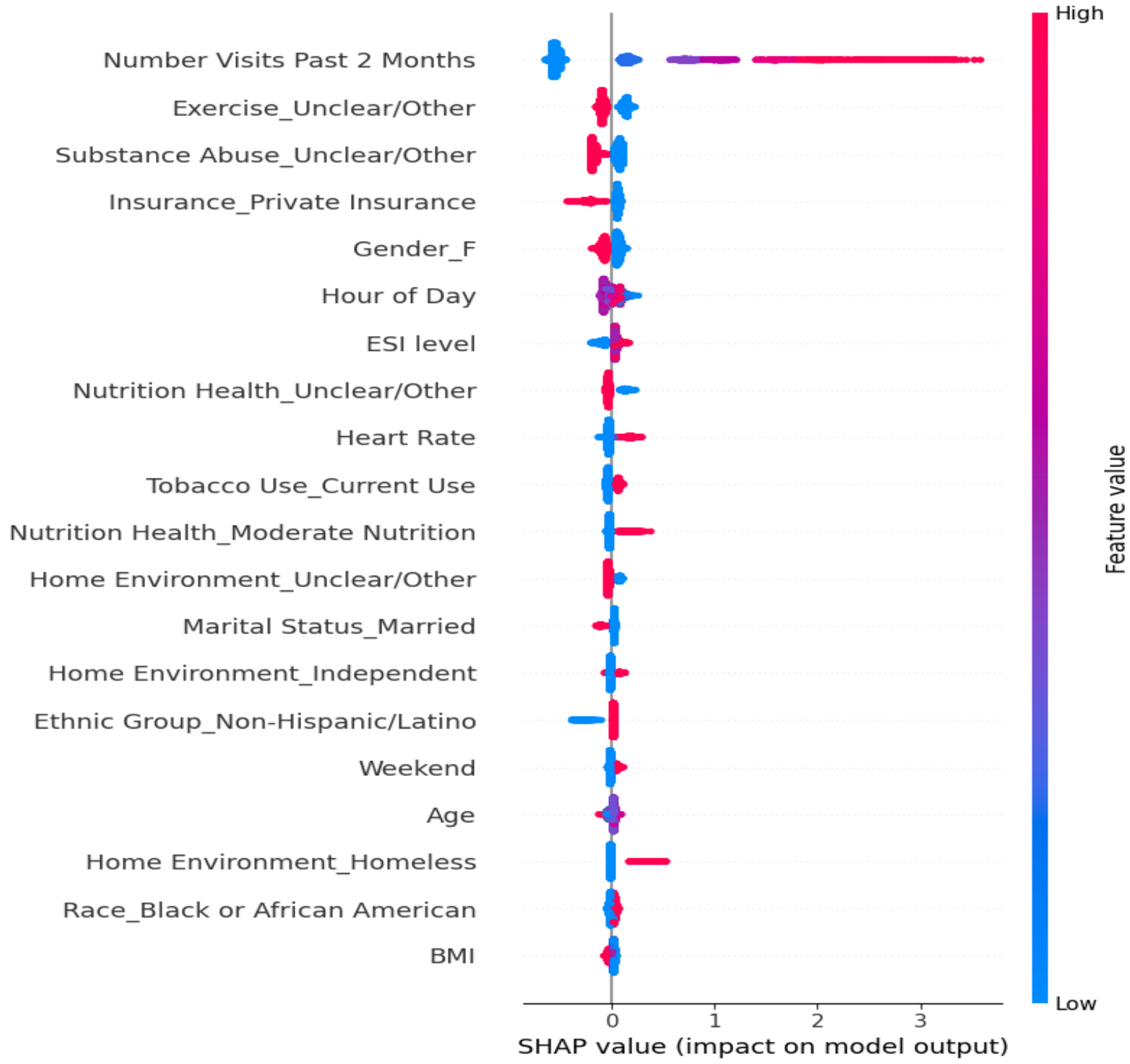

*Figure 2. SHAP Feature Importance in Predicting ED Return*

*Table 6: Comparison of LLM and SHAP-based expandability*

**Risk: High**

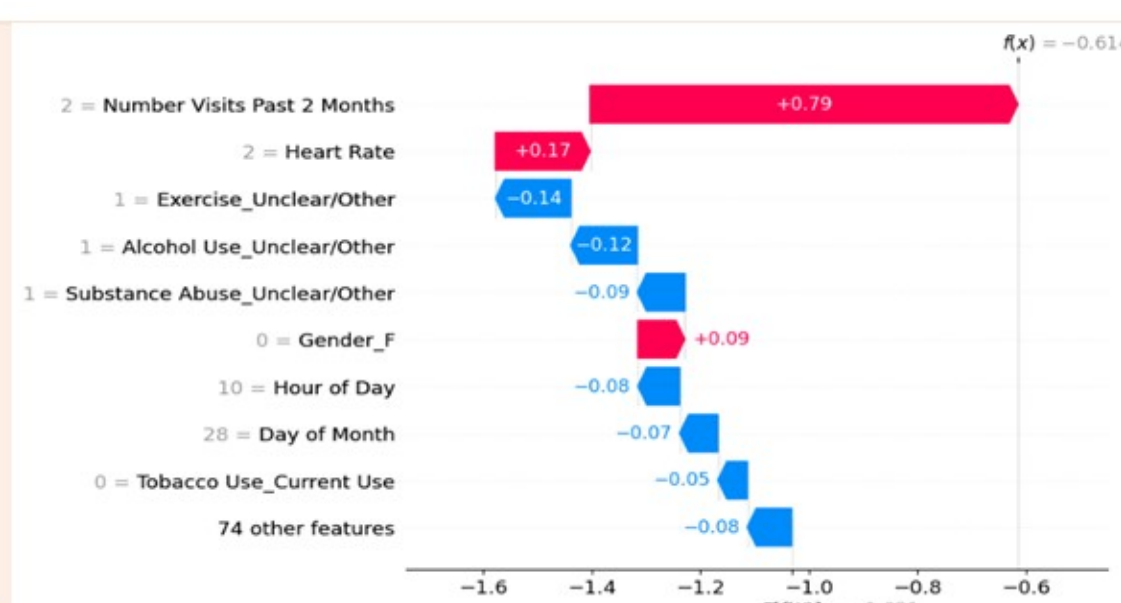

**LLM-Based Explainability:-**

- **Based on the patient's recent history and medical data, they are classified as High risk for a mental health emergency return. The top contributing features include frequent visits in the past two months (SHAP=0.788), tachycardia (SHAP=0.174), and unclear/other exercise habits (SHAP=-0.139). These features indicate increased risk for mental health issues, with the number of visits being a strong predictor of potential emergencies. In comparison to the population, the patient's heart rate is elevated, which can be a sign of stress or anxiety. The unclear/other exercise category may suggest a lack of physical activity, which can also increase mental health risk.**
- **Alcohol use and substance abuse are unclear/other, but their impact on risk is decreased compared to other features (SHAP=-0.123 and -0.087, respectively). While these factors may contribute to overall mental health risk, they are not as significant as the top features in this case.**
- **Overall, the patient's high number of recent visits and tachycardia suggest a higher likelihood of mental health emergencies. Their unclear/other exercise habits and decreased alcohol use/substance abuse may also contribute to their overall risk level. Based on these findings, it is justified to classify the patient as High risk for a mental health emergency return.**

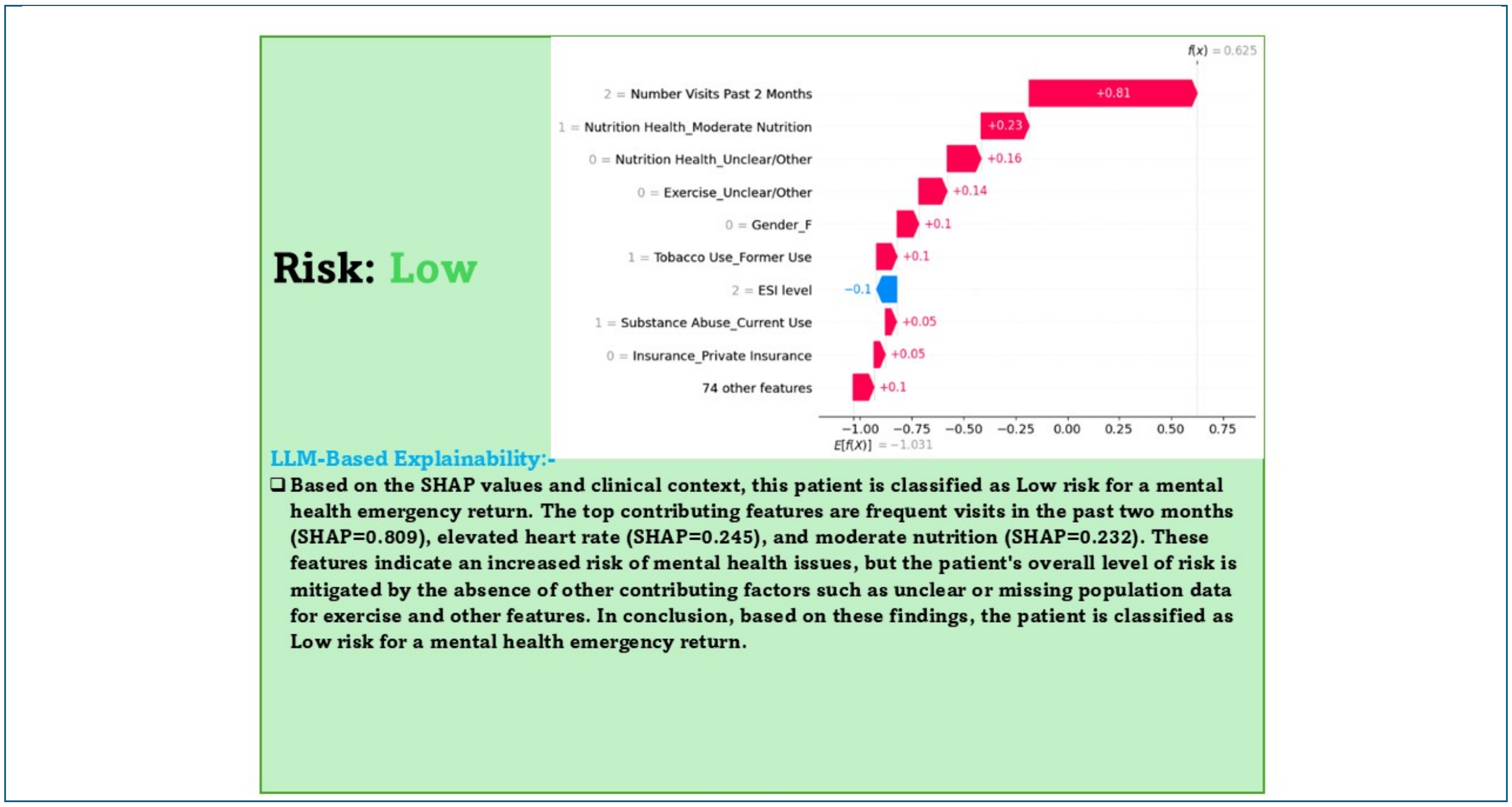

## 5. Discussion

This study introduces a layered clinical AI framework that advances both prediction accuracy and interpretability by integrating structured data, LLM-processed unstructured features, and narrative explanations. The first layer focuses on enriching the input data through LLM-driven processing of free-text fields, including chief complaints and SDoH. These LLM-derived features—such as chief complaints or nuanced social risks—add patient-level context often missed in structured EHR variables. Prior research has highlighted the value of free-text data in capturing clinically relevant information that does not present in structured fields [46, 47]. For instance, in our study, the LLM (Llama 3, 8-billion) with 10-shot learning achieved superior performance in classifying chief complaints, with an accuracy of 0.882 and F1-score of 0.86, markedly outperforming traditional models like XGBoost (Accuracy: 0.59, F1: 0.53). Similarly, in SDoH processing, the same LLM model demonstrated strong classification capabilities with an overall weighted F1-score of 0.96, especially excelling in Alcohol, Tobacco, and Substance Abuse categories (F1: 0.96–0.89). Even in challenging areas like Home Environment and Exercise, the model maintained reasonable accuracy (F1: 0.67–0.70), underscoring its robustness in capturing subtle clinical and behavioral nuances from text. These findings align with prior evidence that transformer-based language models outperform traditional NLP pipelines in extracting contextual signals from EHRs [48, 49], illustrating the value of incorporating LLM-based representations of unstructured data into clinical AI systems to enhance their contextual depth and predictive utility.

In the second layer, the enriched feature set—comprising both structured and LLM-processed inputs—is used to train ML models, particularly XGBoost, to predict the risk of MH-related ED returns. The inclusion of LLM-derived features significantly enhanced model performance, with the area under the AUC improving from 0.74 to 0.76 and AUC-PR rising from 0.58 to 0.61. This gain demonstrates the added predictive value of incorporating semantically enriched information into traditional structured models. These findings align with recent work showing that hybrid models combining structured and unstructured data outperform structured-only models in predicting outcomes such as hospital readmissions and clinical deterioration [50, 51]. Importantly, the hybrid model in this study achieved its performance improvements while maintaining generalizability and requiring only modest increases in computational complexity—an essential consideration for real-time deployment in clinical settings [51]. Such results suggest that integrating LLM-derived context from clinical narratives can bridge gaps in structured data and improve prediction of complex, multifactorial outcomes like ED return for MH patients.

The third layer of the framework emphasizes explainability. Rather than relying solely on SHAP visualizations, which may be difficult for clinicians to interpret, this study employed LLMs to generate structured, narrative explanations grounded in SHAP values and population statistics. These LLM-based narratives systematically convey key contributing features, contextualize patient-specific values against normative data, and clarify whether each factor elevates or reduces risk. By rendering complex algorithmic outputs into clear clinical language, the approach addresses one of the most cited barriers to AI adoption in medicine—model opacity [52]. Similar to efforts by Ribeiro et al. [53] and Lundberg et al. [54] to bridge human-AI understanding through local explanations, our method advances usability by embedding explanations into clinical logic. This design enables actionable, trustworthy insights at the point of care. Collectively, the three-layered framework—data enrichment, predictive modeling, and LLM-based explanation—represents a comprehensive decision-support pipeline that tackles key challenges in clinical AI, including fragmented data inputs, interpretability concerns, and provider trust.

Despite the promising outcomes, this study has several limitations that should be considered. First, although the explainability framework achieved high accuracy in generating clinical narratives, its actual influence on clinician trust and decision-making was not formally evaluated. Second, the study was conducted within a single academic medical center, which may restrict the generalizability of the findings to institutions with different patient demographics, documentation styles, and clinical practices. Third, while the layered framework integrates LLMs to enhance interpretability and data enrichment, the associated computational demands and latency may pose barriers to real-time clinical deployment, particularly in resource-limited settings. These limitations point to critical directions for future research. Subsequent work should empirically assess how LLM-generated explanations influence clinician behavior, diagnostic accuracy, and trust in AI-assisted decision-making. Multicenter validation is necessary to ensure that the framework performs reliably across diverse healthcare environments. Additionally, optimization of model efficiency and infrastructure is needed to enable real-time implementation within electronic health records. Finally, expanding the use of LLMs to extract context from a broader range of clinical narratives—including progress notes, discharge summaries, and social histories—may further improve model relevance and explainability. These steps are essential to advancing trustworthy, interpretable, and scalable AI solutions for clinical care.

## 6. Conclusion

This study advances the field of clinical ML by introducing a layered framework that integrates LLM-based feature extraction, predictive modeling, and explainability to simultaneously enhance accuracy and interpretability. The approach demonstrates that unstructured clinical narratives—when processed via few-shot LLMs—can enrich structured data inputs, improving prediction of MH-related ED returns with minimal labeled data. The final layer leverages LLMs to translate SHAP values into clinician-friendly explanations, bridging the gap between ML outputs and clinical reasoning. This pipeline achieved high performance and interpretive reliability, suggesting that advanced AI tools can be deployed in clinical settings without compromising transparency or usability. Future work should focus on evaluating clinician trust in LLM-based explanations, validating the framework across diverse healthcare systems, and optimizing computational performance for real-time deployment.

## 7. Data availability statement

Data related to this paper are available with the authors and will be available upon reasonable request.

## 8. Funding Statement

This study did not receive any funding.

## 9. Institutional Review Board (IRB) Statement

The Institutional Review Board (IRB) of the University of Alabama at Birmingham (UAB) determined that this study (IRB # IRB-300008858) does not meet the criteria for human subjects research and therefore does not require ethical approval.

# 10. Appendix A: Prompt Templates Used for LLM Classification

## 10.1 A.1 Chief Complaint Classification Prompt

You are a medical classification assistant. Classify the following emergency department chief complaints into one of five categories: Pain, Psychiatric, Injury, Infection, or Unclear.

Examples:
1. Chief Complaint: "I can't stop vomiting." → Category: Infection
2. Chief Complaint: "Severe back pain after lifting boxes." → Category: Pain
3. Chief Complaint: "Hearing voices and suicidal thoughts." → Category: Psychiatric
4. Chief Complaint: "Cut hand with a kitchen knife." → Category: Injury
5. Chief Complaint: "Weakness for the past week, unknown cause." → Category: Unclear

Now classify the following:

Chief Complaint: "[NEW_CHIEF_COMPLAINT]"
→ Category:

## 10.2 A.2 Alcohol Use Classification Prompt

You are a clinical classification assistant. Classify patient-reported alcohol use into one of the following categories:
- No Alcohol Use
- Current Alcohol Use
- Past Alcohol Use
- Occasional Use
- Recovering
- Unclear/Other

Examples:
1. Input: "No alcohol ever" → Category: No Alcohol Use
2. Input: "Drinks socially, rarely" → Category: Occasional Use
3. Input: "History of alcohol abuse, now sober" → Category: Recovering
4. Input: "Used to drink, quit 5 years ago" → Category: Past Alcohol Use
5. Input: "Drinks 3-4 times/week" → Category: Current Alcohol Use
6. Input: "No mention" → Category: Unclear/Other

Now classify:
Input: "[ALCOHOL_TEXT]"
→ Category:

## 10.3 A.3 Nutrition Health Classification Prompt

You are a clinical assistant classifying nutrition-related responses. Use the following categories:
- Balanced Diet
- Unhealthy Diet
- Irregular Eating Habits
- Malnutrition Risk

- Unknown/Other

Examples:
1. Input: "Eats fast food every day" → Category: Unhealthy Diet
2. Input: "Three meals a day, includes vegetables" → Category: Balanced Diet
3. Input: "Sometimes skips meals" → Category: Irregular Eating Habits
4. Input: "Underweight and reports poor appetite" → Category: Malnutrition Risk
5. Input: "No data provided" → Category: Unknown/Other

Now classify:
Input: "[NUTRITION_TEXT]"
→ Category:

### 10.4 A.4 Tobacco Use Classification Prompt

You are a clinical assistant. Classify tobacco use into one of the following:
- Never Smoked
- Current Smoker
- Former Smoker
- Occasional Smoker
- Vaping Only
- Unknown/Other

Examples:
1. Input: "Smokes daily, about a pack" → Category: Current Smoker
2. Input: "Quit 2 years ago" → Category: Former Smoker
3. Input: "Never smoked" → Category: Never Smoked
4. Input: "Uses e-cigarettes occasionally" → Category: Vaping Only
5. Input: "No clear response" → Category: Unknown/Other

Now classify:
Input: "[TOBACCO_TEXT]"
→ Category:

### 10.5 A.5 Substance Abuse Classification Prompt

You are a clinical classification assistant. Categorize substance use into:
- No Substance Use
- Current Use
- Past Use
- In Recovery
- At Risk
- Unclear/Other

Examples:
1. Input: "Currently using methamphetamines" → Category: Current Use
2. Input: "Recovering from opioid addiction" → Category: In Recovery
3. Input: "Never used drugs" → Category: No Substance Use
4. Input: "Occasional marijuana use in the past" → Category: Past Use

5. Input: "History of use, unsure if still using" → Category: Unclear/Other

Now classify:
Input: "[SUBSTANCE_TEXT]"
→ Category:

### 10.6 A.6 Exercise Classification Prompt

Classify the patient's physical activity level into:
- Regular Exercise
- Sedentary Lifestyle
- Occasional Activity
- Limited Mobility
- Unclear/Other

Examples:
1. Input: "Walks daily for 30 minutes" → Category: Regular Exercise
2. Input: "No time for exercise" → Category: Sedentary Lifestyle
3. Input: "Exercises once or twice a month" → Category: Occasional Activity
4. Input: "Wheelchair bound" → Category: Limited Mobility
5. Input: "Not specified" → Category: Unclear/Other

Now classify:
Input: "[EXERCISE_TEXT]"
→ Category:

### 10.7 A.7 Housing Environment Classification Prompt

You are a clinical assistant classifying a patient's housing situation:
- Stable Housing
- Unstable Housing
- Homeless
- Transitional Housing
- Lives With Others
- Unclear/Other

Examples:
1. Input: "Has own apartment" → Category: Stable Housing
2. Input: "Living in a shelter" → Category: Homeless
3. Input: "Staying temporarily with friends" → Category: Transitional Housing
4. Input: "Lives with parents" → Category: Lives With Others
5. Input: "No mention of housing" → Category: Unclear/Other

Now classify:
Input: "[HOUSING_TEXT]"
→ Category:

### 10.8 A.8 Sexual Orientation Classification Prompt

Classify the patient's sexual orientation into:
- Heterosexual
- Homosexual
- Bisexual
- Other Identity
- Declined to Answer
- Unclear/Unknown

Examples:
1. Input: "Straight" → Category: Heterosexual
2. Input: "Gay man" → Category: Homosexual
3. Input: "Bisexual" → Category: Bisexual
4. Input: "Prefers not to say" → Category: Declined to Answer
5. Input: "Queer" → Category: Other Identity
6. Input: "Not clear from note" → Category: Unclear/Unknown

Now classify:
Input: "[SEXUAL_ORIENTATION_TEXT]"
→ Category:

### 10.9 A.9 Patient Risk Analysis Explanation Prompt

You are a medical risk analyst. Write a clear and concise explanation (maximum 200 words) for why this patient is classified as {risk_level} risk for a mental health emergency return. Use plain, clinically relevant language that is easy to understand.

SHAP values indicate the impact of each feature on the model's risk prediction:

- A positive SHAP value means the feature increases the patient's risk.

- A negative SHAP value means the feature decreases the patient's risk.

Below are the top 10 features most responsible for this patient's classification:

{features}

Population-level context for these features:

{population_stats}

Your explanation should follow this structure:

1. Start with a brief statement summarizing the patient's overall risk level and top contributing features.

2. For each feature, explain how it affects risk using clinical terms (e.g., "frequent visits", "elevated heart rate"), and compare to population values if available.

- Do not include SHAP values mid-sentence; instead, include them at the end of each item in parentheses (e.g., SHAP=0.245).

3. Group features with unclear or missing population data together in one paragraph.

4. Conclude with a one-sentence summary justifying the patient's overall risk classification based on the data.

Avoid using symbols or technical jargon (e.g., no arrows like ↑ or ↓, no equations). Do not include the patient index.

Analysis (max 200 words):